\pdfoutput=1

\documentclass[11pt]{article}

\usepackage{emnlp2021}
\usepackage{times}
\usepackage{latexsym}
\usepackage[T1]{fontenc}
\usepackage[utf8]{inputenc}
\usepackage{soul}
\usepackage{url}
\usepackage{xcolor}
\usepackage{graphicx}
\usepackage{amsmath}
\usepackage{amsthm}
\usepackage{booktabs}
\usepackage{algorithm}
\usepackage{algorithmic}
\usepackage{todonotes}
\usepackage{subcaption}
\usepackage{multirow}
\usepackage{IEEEtrantools}
\usepackage{microtype}
\usepackage{lscape, afterpage}
\usepackage{pifont}
\newcommand{\cmark}{\ding{51}}%
\newcommand{\xmark}{\ding{55}}%
\usepackage{tabularx, makecell}%

\title{Does External Knowledge Help Explainable Natural Language Inference? \\ Automatic Evaluation vs. Human Ratings}

\author{Hendrik Schuff$^{1,2}$ \and Hsiu-Yu Yang$^{1,2}$ \and Heike Adel$^1$ \and Ngoc Thang Vu$^2$\\
    $^1$Bosch Center for Artificial Intelligence, Renningen, Germany\\
    $^2$Institut für Maschinelle Sprachverarbeitung, University of Stuttgart\\
    \texttt{\{Hendrik.Schuff,Heike.Adel\}@de.bosch.com}\\
    \texttt{\{Hsiu-Yu.Yang,Thang.Vu\}@ims.uni-stuttgart.de}
}

\begin{document}
\maketitle
\begin{abstract}
Natural language inference (NLI) requires models to learn and apply commonsense knowledge.
These reasoning abilities are particularly important for explainable NLI systems that generate a natural language explanation in addition to their label prediction.
The integration of external knowledge has been shown to improve NLI systems, here we investigate whether it can also improve their explanation capabilities.
For this, we investigate different sources of external knowledge and evaluate the performance of our models on in-domain data as well as on special transfer datasets that are designed to assess fine-grained reasoning capabilities.
We find that different sources of knowledge have a different effect on reasoning abilities, for example, implicit knowledge stored in language models can hinder reasoning on numbers and negations.
Finally, we conduct the largest and most fine-grained explainable NLI crowdsourcing study to date.
It reveals that even large differences in automatic performance scores do neither reflect in human ratings of label, explanation, commonsense nor grammar correctness.
\end{abstract}
\section{Introduction}
\begin{figure}[t]
    \centering
    \includegraphics[width=\columnwidth]{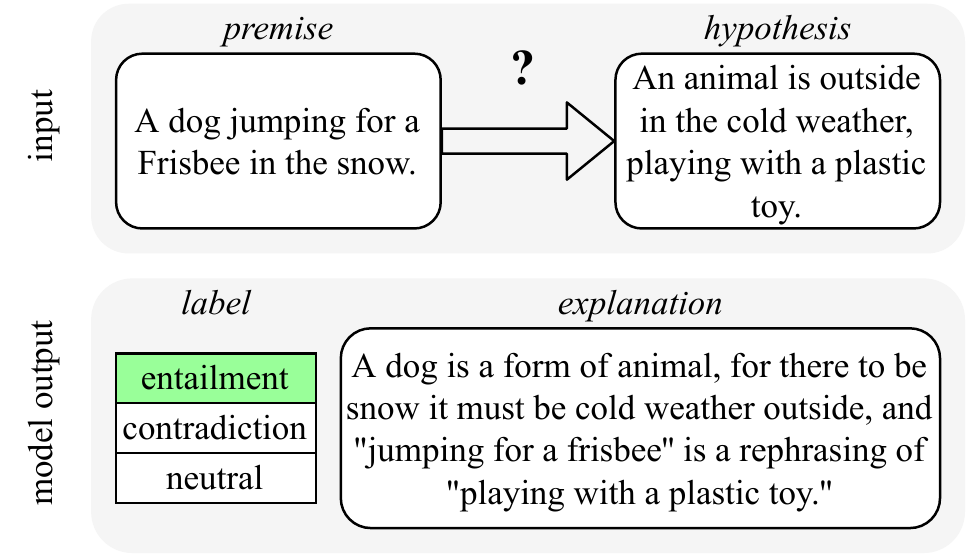}
    \caption{NLI example instance from e-SNLI \protect\cite{camburu_e-snli_2018}. The system needs to include commonsense knowledge, such as ``snow $\rightarrow$ cold weather''.}
    \label{fig:nli_example}
\end{figure}

Natural language inference (NLI) is closely related to real-world applications, such as fact checking.
Given two sentences (premise and hypothesis), the task is to decide whether
(a) the first sentence entails the second sentence, (b) the two sentences contradict each other or (c) they have a neutral relation.
Figure~\ref{fig:nli_example} shows an example for an entailment relation.
Solving the task requires models to not only reason over the provided information but also to link it with commonsense knowledge.

As for other natural language tasks, state-of-the-art NLI systems rely on deep neural architectures which do not easily expose their inner workings.
However, following a model's reasoning {process is valuable to machine learning engineers as well as end-users. The former can use the insights to improve models and the latter can base their decision on 
them whether to trust the system or not. 
One approach to gain insight into a system is to train it to generate 
explanations as an additional output \citep{camburu_e-snli_2018,atanasova-etal-2020-generating-fact}.
Such self-explaining models are particularly interesting for NLI because the explanation can indicate the commonsense knowledge which was utilized during prediction.
The integration of external knowledge was shown to improve NLI systems \citep{jijkoun_recognizing_2005,chen-etal-2018-neural,li_several_2019, Faldu2021KiBert}.
However, the following question remains:  \textbf{Does the positive effect of external knowledge on the inference ability transfer to the generation of explanations? (R1)}
Figure~\ref{fig:nli_example} shows an NLI example for which external knowledge potentially helps to infer the correct label and explanation.
In the example, the system needs to link \textit{dog} to \textit{animal}, \textit{jumping for a Frisbee} to \textit{playing}, \textit{Frisbee} to \textit{plastic toy} and \textit{snow} to \textit{outside} as well as to \textit{cold weather}.
The predicted explanation needs to explicitly state this reasoning chain and thus would be expected to benefit from external knowledge.

Especially recently, pre-trained language models, such as BERT \citep{devlin-etal-2019-bert} or GPT \citep{radford_language_2019}, became popular.
It was shown that they are able to learn and store commonsense knowledge implicitly \citep{petroni-etal-2019-language}.
As a result, an open question is:
\textbf{How effective is the implicit commonsense knowledge of language models compared to symbolic sources of knowledge, such as knowledge base triplets? (RQ2)}

To evaluate NLI models, mainly automatic measures, such as accuracy, are used. However, model weaknesses can stay unnoticed using automatic scores alone. Moreover, \citet{schuff-etal-2020-f1} showed that automatic scores are not necessarily correlated to human-perceived model quality. Thus,
human evaluation is a crucial step in the development of user-centered explainable AI systems.
Therefore, we ask the question: \textbf{How do humans perceive explanation quality of state-of-the-art natural language inference models? (R3)}

In this paper, we investigate the three previously mentioned research questions R1--R3. 
To answer them, we analyze the impact of external knowledge from multiple sources, such as knowledge graphs, embeddings and language models and propose novel architectures to include and combine them into explainable NLI systems.
Further, we conduct an extensive automatic analysis as well as a user study.
To the best of our knowledge, our study exceeds previous human evaluations of explainable NLI models regarding the number of participants as well as the variety of rated explanation criteria.

For R1, we find that the positive effect of external knowledge on label accuracy in the standard NLI setting can also be observed in the explainable NLI setting
and external knowledge can improve the BLEU scores of the generated explanations.
In regard of R2, we observe that pre-trained language models are the most promising source of commonsense knowledge but at the same time identify weaknesses with respect to negations and numerical reasoning abilities which, however, can be mitigated through combination with additional knowledge sources.
Despite the improvements in accuracy, BLEU or BLEURT scores, our user study shows for R3 that these do not reflect in human ratings of explanation correctness, commonsense inclusion or grammar and label correctness. 

Our results urge caution to solely rely on automatic scores for explainability. Therefore, we expect our paper to motivate the development of dedicated evaluation tasks and scores and further emphasize the importance of the user within explainable AI. To facilitate future work, we release our model's predictions as well as the 
crowdsourced human ratings.\footnote{\url{https://github.com/boschresearch/external-knowledge-explainable-nli}}
\section{Related Work}
\subsection{External Knowledge for NLI}
External knowledge was shown to help across a wide variety of NLP tasks \citep{shi-etal-2016-knowledge,seyler-etal-2018-study,pan_improving_2019,lin_kagnet_2019}.
While early sources for external knowledge are WordNet and NomBank \citep{jijkoun_recognizing_2005,maccartney-etal-2008-phrase}, today a large variety of possibilities exist: From COMET \citep{bosselut-etal-2019-comet} over ConceptNet \citep{speer_conceptnet_2017} to language models.
\citet{chen-etal-2018-neural} show that enriching an NLI system with external lexical-level semantic knowledge increases accuracy scores on SNLI and enhances transfer to MultiNLI.
\citet{wang_improving_2019} show the potential of knowledge from ConceptNet for NLI systems.
\citet{li_several_2019} find that external knowledge from pre-training helps NLI and suggest to combine it with external knowledge from human-curated resources.
\citet{li_knowledge_2019} propose knowledge-enhanced attention modifications for Transformers and decomposable methods and show that their methods improve model robustness.
\citet{Faldu2021KiBert} extend BERT by extracting entities from the input text and adding their projected KG embeddings derived from ConceptNet and WordNet as sequential input to a modified BERT layer.
\citet{bauer_ernie-nli_2021} present ERNIE-NLI, a modified ERNIE \cite{zhang_ernie_2019} model using NLI-specific knowledge embeddings and find that it improves performance over a non-adapted ERNIE model using general-domain TransE embeddings.
To compare different possibilities of integrating external knowledge, we propose various models in this paper.
Further, we address the question whether external knowledge also improves explanation generation.

\subsection{Explainable NLI}
The task of explainable NLI consists of (i) predicting the correct entailment label and (ii) providing an explanation that allows the user to assess the model's reasoning.
In general, such explanation can take various forms, such as weights and gradients over the input \citep{simonyan_deep_2014,ribeiro_why_2016,lundberg_unified_2017} and text spans or snippets from the input or external text \citep{zaidan-eisner-2008-modeling,lei-etal-2016-rationalizing,yang-etal-2018-hotpotqa}.
Beyond that, there exists various resources and approaches designed to generate textual explanations.
\citet{rajani-etal-2019-explain} present a dataset that contains free-text explanations for multiple-choice commonsense reasoning and \citet{bhagavatula_abductive_2020} provide a dataset for abductive multiple choice answering as well as abductive NLG.
\citet{camburu_e-snli_2018} provide the e-SNLI dataset, which adds free-text explanations as an additional layer on the SNLI dataset \citep{bowman-etal-2015-large}.
As numerous models with and without external knowledge have been developed on the SNLI dataset, we use its explainable extension e-SNLI to conduct our analysis and train our models on.
Various models have been proposed on e-SNLI including systems based on alignment \citep{swanson-etal-2020-rationalizing}, label-specific explanation generators \citep{kumar-talukdar-2020-nile} and fine-tuned text-to-text models \citep{narang_wt5_2020}.
In contrast to those, our focus is not on proposing a new architecture or paradigm to develop a high-scoring system.
Much more, we seek to conduct a broad comparison across knowledge sources and isolate their effect on automatic scores as well as human perception.

\subsection{Evaluation and Human Ratings}
Explainable NLI system performance is typically scored using (i) accuracy with respect to annotated gold labels on a reference dataset and (ii) BLEU scores \cite{papineni-etal-2002-bleu} between the generated explanations and the ground truth explanations \cite{camburu_e-snli_2018,kumar-talukdar-2020-nile,narang_wt5_2020}.
BLEU scores can only quantify explanation quality loosely \cite{narang_wt5_2020}. 
Therefore, previous work evaluates explanation quality either by manual annotation \cite{camburu_e-snli_2018,kumar-talukdar-2020-nile} or crowdsourcing \cite{narang_wt5_2020}.
However, previous human evaluations regarding explainable NLI are limited to assess label and/or explanation correctness.
In contrast, we additionally evaluate commonsense inclusion as well as grammatical correctness of explanations.
As \citet{clinciu-etal-2021-study} find automatic BLEURT scores to have distinctly stronger correlations to human ratings of generated textual explanations than BLEU, we investigate whether BLEURT is a viable replacement for a user study.
\section{Method}
In the following, we describe our base model and then present the models we analyze.

\subsection{Base Model}
We combine a state-of-the-art attention-based inference model with an explainable NLI model that predicts entailment labels and generates explanations.
In particular, we use the encoder part of the enhanced sequential inference model (ESIM), which has a cross-attention layer to capture relevant semantics between premise and hypothesis \citep{chen-etal-2017-enhanced} and the prediction part of the \textsc{pred-expl} model of \citet{camburu_e-snli_2018}.
We represent the input sentences with BERT embeddings \citep{devlin-etal-2019-bert} which we fine-tune on the SNLI dataset.\footnote{We pass inputs of the form \textit{[CLS] premise [SEP] hypothesis} to BERT and use a softmax layer on top of the CLS token's embedding to predict the entailment label and fine-tune the model for up to 2 epochs.}
Throughout the paper, we refer to this model as \textsc{Vanilla}.

\subsection{Integration of Knowledge Sources}
External knowledge can be found in various formats.
We aim at covering a possibly broad variety and focus on state-of-the-art sources and methods.
We include the natural language knowledge base COMET \citep{bosselut-etal-2019-comet}, the ConceptNet Numberbatch embeddings \citep{speer_conceptnet_2017} and the GPT-2 language model \citep{radford_language_2019}.

\subsubsection{Background Knowledge from COMET}
As our example in Figure~\ref{fig:nli_example} shows, resolving natural language entailment can require reasoning over multiple concepts and relations, such as inferring \textit{cold weather} and \textit{outside} from \textit{snow}.
We seek to facilitate this resolvement by providing the model with related words (and phrases) that can be seen as a natural language extension of premise and hypothesis. We use the COMmonsEnse Transformers (COMET) \cite{bosselut-etal-2019-comet} as a natural language knowledge base to query background knowledge for premise and hypothesis.
COMET is based on a transformer language model that is fine-tuned on a knowledge base completion task on ConceptNet. 
Given an input sentence and a ConceptNet relation, it generates a phrase to complete the object in a knowledge statement expressed in the (subject, relation, object) format.
Instead of feeding in the whole premise and hypothesis, we find that chunking them into noun and verb sub-phrases based on POS tags patterns yields better object phrase generations.\footnote{We manually find that feeding in the whole sentence predominantly relates the output to the last tokens of the sentence and fails to include information from tokens earlier in the sentence.}
Thus, for each sentence (premise/hypothesis) we generate \#chunks $\times$ \#relations object phrases.\footnote{We consider the relations AtLocation, CapableOf, DefinedAs, HasA, HasProperty, HasSubevent, InheritsFrom, InstanceOf, IsA, LocatedNear,  MadeOf, PartOf, SymbolOf, UsedFor and LocationOfAction.} 

Afterwards, we embed each object phrase (with the respective relation string prependend) with Sentence-BERT \citep{reimers-gurevych-2019-sentence} and quantify its similarity to the embedding of the source sentence using cosine similarity.
For each relation, we keep the object phrase with the highest similarity score.

 Given the relation \textit{HasA}
 and the chunked sentence
 \textit{The dog} | \textit{is walking in the snow},
 for example, COMET will generate \textit{bone} and \textit{effect of freeze} for the two sub-phrases, respectively. We only preserve the object phrase \textit{effect of freeze} as it has a higher similarity to the source sentence.

To condense the object phrases into a fixed-length vector representation, we average the respective Sentence-BERT embeddings.
This procedure yields one vector representing the background knowledge regarding the premise and one regarding the hypothesis.
We combine them with the local inference vector representation \cite{chen-etal-2017-enhanced}. 
Following \citet{camburu_e-snli_2018}, this vector is passed to the label prediction module as well as the explanation decoder.
We refer to this model as \textsc{comet}.

\subsubsection{Modified Attention with ConceptNet}
Following \citet{li-srikumar-2019-augmenting}, we use knowledge-driven rules to modify the attention weights within the cross-attention layer between premise and hypothesis in the encoder.
This enforces the attention mechanism to align word pairs $p_{i}$ and $h_{j}$ from premise and hypothesis based on world knowledge.
The rules proposed by \citet{li-srikumar-2019-augmenting} are shown in Equation \ref{eqn:R1} and \ref{eqn:R2}. In $R_1$, the antecedent $K_{p_{i},h_{j}}$ indicates that a word pair $p_{i}$ and $h_{j}$ is of a certain relation within ConceptNet.
If the condition of the antecedent is true, the consequent $ A_{p_{i},h_{j}}'$ that aligns the word pair follows.
$R_2$ is a relatively conservative rule that additionally takes the model's own decision into account.
The antecedent $K_{p_{i},h_{j}} \land A_{p_{i},h_{j}}$ in $R_2$ is a conjunctive condition that becomes true if a word pair is both in a relation and aligned by a model's original attention. If such a conjunctive condition is true, the word pair must be aligned which results in a new alignment as the consequent $A_{p_{i},h_{j}}'$ indicates.
\begin{align}
        	R_1: K_{p_{i},h_{j}} \to A_{p_{i},h_{j}}'\label{eqn:R1}\\
        	R_2: K_{p_{i},h_{j}} \land A_{p_{i},h_{j}} \to A_{p_{i},h_{j}}'\label{eqn:R2}
\end{align}
Different from the approach of \citet{li-srikumar-2019-augmenting} that checks a word pair's relation in a binary fashion, we hypothesize that knowledge-aware embeddings might capture more fine-grained word relationship that exists in multi-hop relational edges.
Considering \textit{playground} and \textit{playroom}, for example, the former is usually located outdoors whereas the latter is located indoors.
We generalize the binary relational inclusion from \citet{li-srikumar-2019-augmenting} to continuous relation scores.
For this, we replace the binary rule antecedent with the absolute cosine similarity between the ConceptNet Numberbatch \citep{speer_conceptnet_2017} vector representations of $p_{i}$ and $h_{j}$.
We empirically confirm that our continuous formulation outperforms the binary version regarding label accuracy as well as explanation correctness.
In the following, we refer to these modified rules as continuous constraints and use \textsc{cont} to refer to the respective model.

\subsubsection{All-Text Prediction with GPT-2}
Similar to \citet{kumar-talukdar-2020-nile}, we fine-tune a pre-trained GPT-2 language model on the e-SNLI dataset.
In contrast to \citet{kumar-talukdar-2020-nile}, we use a single GPT-2 model to generate explanations for all three entailment labels instead of training a separate model for each of them.
This allows us to directly integrate the label prediction into the language model instead of training an additional model which predicts the label on top of the three explanations.
Therefore, we propose two models, which both are GPT-2-large models, but differ regarding their training setting.
In the label-first setting (\textsc{gpt-lf}), the model is trained on text following the structure \textit{Premise: $<$premise$>$ Hypothesis: $<$hypothesis$>$ [LAB] [label] [EXP] $<$explanation$>$ EOS}.
In the explanation-first setting (\textsc{gpt-ef}) it is trained on text following the structure \textit{Premise: $<$premise$>$ Hypothesis: $<$hypothesis$>$ [EXP] $<$explanation$>$ [LAB] $<$label$>$ EOS}.

\subsection{Combined Models}
\label{sec:combined_models}
\paragraph{COMET and ConceptNet.}
We combine \textsc{comet} with \textsc{cont} to benefit from both integrated background information from COMET and a knowledge-enhanced attention mechanism based on ConceptNet Numberbatch. We expect this to help the model focus on important relations between premise and hypothesis.

\paragraph{Knowledge-Enhanced Ensembles.}
We combine the world knowledge of BERT (\textsc{vanilla}), ConceptNet Numberbatch (\textsc{cont}), COMET (\textsc{comet}) and the combined model \textsc{comet+cont} with the language model abilities of GPT-2 (\textsc{gpt-lf} and \textsc{gpt-ef}).
For this, we propose an ensemble that not merely aggregates label votes, but combines the models with respect to their different strengths.

The label predictions of \textsc{vanilla}, \textsc{cont}, \textsc{comet}, \textsc{comet+cont} as well as \textsc{gpt-lf} are passed to a majority voting.
In the \textit{basic ensemble}, the \textsc{gpt-lf} model is then conditioned on the voted label and generates the final explanation.
We refer to this model as \textsc{ensemble}.

In the \textit{filtered ensemble}, the majority voting only allows models to vote if their generated explanation lets the \textsc{gpt-ef} model predict the same label prediction as the original model.
In other words, we fix the input as well as the generated explanation and only let the \textsc{gpt-ef} model predict the label.
This step can be interpreted as a consistency filter which prevents models from voting if their label prediction does not match their explanation prediction.
In the following, we refer to this model as \textsc{filtered-ens}.
We include a depiction of the model architecture in the appendix.
\section{Automatic Evaluation}\label{sec:auto_eval}
First, we evaluate the discussed knowledge-enhanced models with respect to commonly used scores on e-SNLI and a stress test evaluation.
In addition to our constructed models, we also include \textsc{pred-expl} \citep{camburu_e-snli_2018}, which is basically our \textsc{vanilla} baseline without cross-attention but with GloVe embeddings instead of fine-tuned BERT embeddings.
Further, we include two recent models proposed for e-SNLI: NILE:post-hoc, which is the highest performing model from \citet{kumar-talukdar-2020-nile}, and WT5-11B from \citet{narang_wt5_2020}, which holds the current state-of-the-art performance.
While NILE:post-hoc is based on GPT-2 as well, WT5-11B is a fine-tuned version of the T5 language model \citep{raffel_exploring_2020}. 
We train all non-LM models with five random seeds and report scores of the median model based on label accuracy.
Table~\ref{tab:generated_examples} shows predicted explanations for the subset of models that we investigate within the human evaluation in Section~\ref{sec:human_eval}. Further examples are provided in the appendix.

\subsection{Performance on e-SNLI}
\begin{table}
\centering
\resizebox{\columnwidth}{!}{%
\begin{tabular}{clcccc}
\toprule
Type & Model & Label Acc. & BLEU & BLEURT   \\ \midrule
\multirow{5}{*}{\rotatebox[origin=c]{90}{non-LM}} & \textsc{pred-expl}   & 84.21  & 19.77  & -0.871           \\
 & \textsc{vanilla}                                                              & 89.20  & 19.71  & -0.820           \\
 &\textsc{comet}                                                                 & 88.97  & 18.84  & -0.822           \\
 & \textsc{cont}                                                                 & 89.02  & 20.1  & -0.799            \\
 & \textsc{comet+cont}                                                           & 89.07  & 19.66  & -0.809 \\ \midrule
\multirow{6}{*}{\rotatebox[origin=c]{90}{LM-based}}  & \textsc{gpt-ef}           & 87.89  & 21.70  & -0.624            \\
 & \textsc{gpt-lf}                                                               & 89.70  & 26.90  & -0.577            \\
 & \textsc{ensemble}                                                             & 90.24  & 27.10  & -0.576            \\
 & \textsc{filtered ens}                                                    & 90.24  & 27.09  & -0.577            \\
 \cmidrule(lr){2-6}
 & \textsc{nile:post-hoc}                                                        & 91.49   &   26.26  &  -0.577 &           \\
 & \textsc{wt5-11b}                                                              &\textbf{92.3}~~   &   \textbf{29.01}   & \textbf{-0.511} &           \\ 
 \bottomrule
\end{tabular}
}
\caption{Automatic evaluation metrics on the e-SNLI test set.
Label accuracy quantifies NLI performance. BLEU and BLEURT score the similarity between predicted and ground truth explanation texts.
Higher values are better.
}\label{tab:automatic_scores}
\end{table}
Following prior work on e-SNLI, we report label accuracy as well as BLEU scores \citep{papineni-etal-2002-bleu} for explanations.
We additionally evaluate BLEURT scores \citep{sellam-etal-2020-bleurt}, which is a reference-based learned evaluation metric to model human judgements of text generation.
BLEURT is of particular interest for explanation evaluation as \citet{clinciu-etal-2021-study} compare how various automatic scores such as BLEU, ROUGE and METEOR correlate to human ratings of generated explanations and find that embedding-based methods and particularly BLEURT scores show distinctly higher correlations than, e.g., BLEU.

Table~\ref{tab:automatic_scores} shows the respective scores for all considered models.\footnote{
For NILE:post-hoc \protect\citep{kumar-talukdar-2020-nile} and WT5-11B \protect\citep{narang_wt5_2020} we report the label accuracy from their paper and calculate BLEU/BLEURT scores based on the explanation predictions provided by the authors.
\citet{narang_wt5_2020} calculate BLEU scores using SacreBLEU v1.3. \protect\citep{post_call_2018} leading to a higher reported score of 33.7.
}
The upper block lists models that share or extend the \textsc{pred-expl} architecture.
Compared to \textsc{pred-expl}, the \textsc{vanilla} model achieves a notable increase in label accuracy as well as BLEURT scores.
Surprisingly, \textsc{comet} reduces all scores and even decreases the BLEU score below the \textsc{pred-expl} score.
In contrast, knowledge-enhanced cross attention (\textsc{cont}) improves BLEU and BLEURT scores and reaches a label accuracy close to \textsc{vanilla}.
Combining \textsc{cont} with \textsc{comet} retains the \textsc{cont} label accuracy but again slightly decreases BLEU and BLEURT scores.
The lower block contains models that are or include language models.
All language model-based models increase BLEU and BLEURT scores.
All except \textsc{gpt-ef} outperform all non-language model models.

To analyze whether the performance differences of models can be really attributed to a better reasoning and commonsense knowledge ability instead of merely different model capacity, we next evaluate our models
on the NLI stress test evaluation.

\begin{table}[t!]
    \centering
    \begin{tabular}{p{0.205\columnwidth}p{0.7\columnwidth}}
        \toprule
        Model & Predicted Explanation \\
        \cmidrule(lr){1-1} \cmidrule(lr){2-2}
        \textsc{ground-truth} & a man is either playing the accordion or performs a mime act while happy people pass by or angry people glare at him.\\
        \textsc{vanilla} & a man can not be playing and a mime at the same time\\
        \textsc{comet} & the man is either playing the accordion or a mime\\
        \textsc{cont} & people can not be playing and angry at the same time\\
        \textsc{comet+ cont} & the man can not be playing the accordion and the mime at the same time\\
        \textsc{gpt-lf}& Happy people are not angry people.\\
        \textsc{wt5-11B} & The man cannot be playing the accordion and performing a mime act at the same time.\\
        \bottomrule
    \end{tabular}
    \caption{Explanation predictions of the models used within the human evaluation for the premise  ``\textit{A man on a sidewalk is playing the accordion while happy people pass by}'' and the hypothesis ``\textit{A man on the sidewalk performs a mime act while angry people glare at him}''. All models correctly predict the class \textit{contradiction} but generate different explanations. The predicted explanation of the \textsc{filtered-ens} model is identical to the explanation of the \textsc{gpt-lf} model as \textsc{gpt-lf} is used to predict the ensemble's explanation.}\label{tab:generated_examples}
\end{table}

\subsection{Stress Test Evaluation}\label{sec:stress_tests}
\begin{table*}[htb]
\centering
\resizebox{\textwidth}{!}{%
\begin{tabular}{clccccccc}
\toprule
& & & \multicolumn{2}{c}{Competence Test} & \multicolumn{3}{c}{Distraction Test} & \multicolumn{1}{c}{Noise Test}\\
 \cmidrule(lr){4-5} \cmidrule(lr){6-8} \cmidrule(lr){9-9}
Type &  Model &  Total &  Antonymy &  Numerical &  Word Overlap &  Length Mismatch &  Negation  &  Spelling \\
\midrule
 \multirow{5}{*}{\rotatebox[origin=c]{90}{non-LM}} &      \textsc{pred-expl}   &  48.69 &                36.36 &       36.55 &      47.17 &           53.44 &         45.31 &           52.42 \\
&                  \textsc{vanilla}                                                    &  56.94 &                37.94 &       32.24 &      55.46 &           65.21 &         52.03 &           62.90 \\
&                    \textsc{comet}                                                    &  57.05 &                34.54 &       35.48 &      57.31 &           64.15 &         52.85 &           62.33 \\
&   \textsc{cont}                                                                      &  57.09 &                32.50 &\textbf{40.28}&    52.10 &            64.35 &  \textbf{53.38} &         62.77 \\
&         \textsc{comet+cont}                                                          &  56.26 &                44.43 &       34.16 &      51.34 &           64.39 &         49.36 &           63.03 \\ \midrule
\multirow{4}{*}{\rotatebox[origin=c]{90}{LM-based}} & \textsc{gpt-ef}                  &  52.74 &                51.81 &       31.33 &      55.91 &           60.97 &         38.44 &           58.20 \\
&       \textsc{gpt-lf}                                                                &\textbf{59.28}&  \textbf{54.84}&       28.80 &\textbf{64.06}&\textbf{68.72} &         42.82 &           67.07 \\
&                 \textsc{ensemble}                                                    &  59.19 &                37.97 &       34.03 &      58.13 &           67.45 &         52.51 &           65.92 \\
&        \textsc{filtered-ens}                                                    &  58.99&                 52.53 &       28.54 &      63.70 &           68.02 &         42.18 & \textbf{67.10} \\
\bottomrule
\end{tabular}
}
\caption{
Label accuracies (higher is better) for all categories in the NLI stress test tasks \protect\citep{naik-etal-2018-stress}.
The six rightmost columns show (i) the model's reasoning abilities (competence), (ii) how sensitive it is to lexical distractors (distraction) and (iii) how robust it is against noise from different perturbations (noise).
Each column corresponds to one dataset.
For datasets with matched and mismatched subsets, we report the accuracy over all labels within the group. Similarly, the total accuracy is calculated over all labels.
}\label{tab:stress_tests_detailed}
\end{table*}
Table~\ref{tab:stress_tests_detailed} shows the results of our models on the NLI stress test evaluation proposed by \citet{naik-etal-2018-stress}.
The dataset contains multiple subsets of which each subset is used to evaluate the robustness of the system against a specific type of perturbation, e.g., spelling errors, negations, numerical reasoning and more.
On average, all models distinctly improve performance compared to the \textsc{pred-expl} baseline.
With respect to \textsc{vanilla}, all models except \textsc{gpt-ef} improve average performance.
Further, both \textsc{comet} and \textsc{cont} improve average label accuracy, while their combination decreases performance.
Surprisingly, \textsc{gpt-lf} outperforms the ensemble methods on average.
While \textsc{comet+cont} reaches the best performance in terms of e-SNLI label accuracies, it performs worst on the stress tests.
The same effect can be observed for the \textsc{filtered-ens}.
While it reaches top performance for the spelling error test, its performance drops for numerical reasoning, where it performs worse than any other model.
These results show that combining different knowledge sources does not result in a consistent combination of their weaknesses and strengths.
Instead, the sources of external knowledge have to be carefully adjusted to the target domain and our results paint a rather pessimistic picture regarding a cure-all solution. 
Further, a model's reasoning capabilities have to be assessed in detail as evaluation across different reasoning types easily masks model weaknesses.

Finally, we assess whether language models reach their higher performance due to better reasoning:
For most of the assessed reasoning types --- with exception of numerical reasoning and negation --- the best non-ensemble model in fact is \textsc{gpt-lf}.
Also, \textsc{gpt-lf} reaches the highest accuracy on average.
Therefore one could generally recommend to include external knowledge in form of a pre-trained language model as the foremost option.
However, our results also show that language models are not necessarily the best choice for all reasoning needs and can, e.g., severely decrease performance for numerical reasoning and negations, where models based on language models perform worse than all other models.

\section{Human Evaluation}\label{sec:human_eval}
While automatic scores, such as BLEU, provide a valuable starting point for evaluating explanations, they fall short in capturing the model's real explanation capabilities.
We, therefore, conduct a large-scale crowdsourcing study to complement our automatic evaluations on e-SNLI and the stress tests.
Following related work \citep{narang_wt5_2020}, we assess explanation quality based on ratings from crowdworkers on Mechanical Turk.
While previous work limited evaluation to rating explanation correctness, we additionally ask participants to provide fine-grained ratings of commonsense inclusion and grammatical correctness.
A screenshot of the interface is shown in the appendix.
We release the full data of our study.

\subsection{Conditions}
In order to evaluate effects across the discussed sources of external knowledge, we include seven models in our human evaluation: \textsc{vanilla}, \textsc{comet}, \textsc{cont}, \textsc{comet+cont}, \textsc{gpt-lf}, \textsc{filtered-ens} and \textsc{wt5-11B}.
Additionally we evaluate the e-SNLI ground truth labels and explanations.
Table~\ref{tab:generated_examples} displays the different explanations the models predict for an exemplary input as well as its ground truth explanation annotation.

\subsection{Dependent Variables}
We evaluate the models' predicted labels and explanations along four self-reported dimensions.
\paragraph{Label Correctness.}
Following \citet{kumar-talukdar-2020-nile} and \citet{narang_wt5_2020}, we ask participants to rate if the predicted label is correct.

\paragraph{Explanation Correctness.}
Similar to \citet{camburu_e-snli_2018}, \citet{kumar-talukdar-2020-nile} and \citet{narang_wt5_2020}, we collect subjective yes/no explanation correctness ratings.

\paragraph{Grammatical Correctness.}
We ask participants to rate if the generated explanation is grammatical.

\paragraph{Commonsense Inclusion.}
We ask participants whether the explanation includes commonsense knowledge that is needed to answer the question.
We collect responses on an item with the options \textit{yes}, \textit{no} and \textit{no need}.

\subsection{Study Design}\label{sec:study_design}
In order to evaluate the effect of the level of required external knowledge, we compile, like \citet{kumar-talukdar-2020-nile} and \citet{narang_wt5_2020}, a set of 100 premise-hypothesis pairs.
In contrast to them, we compose the 100 pairs to contain 50 pairs that require a low level of external knowledge and 50 pairs that require a high level.
To gather pairs of both categories, we let two annotators rate 250 premise-hypothesis pairs from the e-SNLI test set.
We sample 50 pairs per level of external knowledge from the 179 pairs on which the annotators agree.
We provide details on the annotation criteria in the appendix.
During the study, we, like \citet{narang_wt5_2020}, collect 5 crowdsourced ratings for each condition and for each of the 10 input pairs per batch, i.e., 500 ratings per model and a total of 4000 ratings for each variable.
We provide ratings of exemplary model predictions in the appendix.

\subsection{Analysis}
We collect responses from 290 crowdworkers and discard those that were entered in less then 5 minutes (31\%) as this might indicate arbitrary answer selection.
Note that the repeated measures design of our study possibly introduces inter-dependencies within ratings as, e.g., certain participants can have a tendency to rate explanations as correct more often than others or a certain question might elicit more label correctness ratings.
Thus, we use generalized linear mixed models (GLMM) to account for the potentially confounding variables (worker ID, question ID and level of required commonsense knowledge).
As our response variables are binary,\footnote{We do not consider ``no need'' commonsense ratings during the respective model estimation.} we use binomial GLMMs.
We include fixed effects (model and commonsense level) as well as random intercepts (worker and question IDs).
Figure~\ref{fig:effect_displays} shows effect displays for the collected ratings in relation to the predictor \textit{model type}.

We conduct likelihood ratio tests between the full model and the model without the evaluated predictor to test the effects of \textit{model type} and \textit{commonsense level} on all four rating variables.
As \textit{model type} contains more than two factors, we additionally conduct single-step corrected Tukey HSD post-hoc tests for all four variables.

\begin{figure*}[t]
    \centering
    \begin{subfigure}[t]{.5\textwidth}
        \centering
        \includegraphics[width=\textwidth]{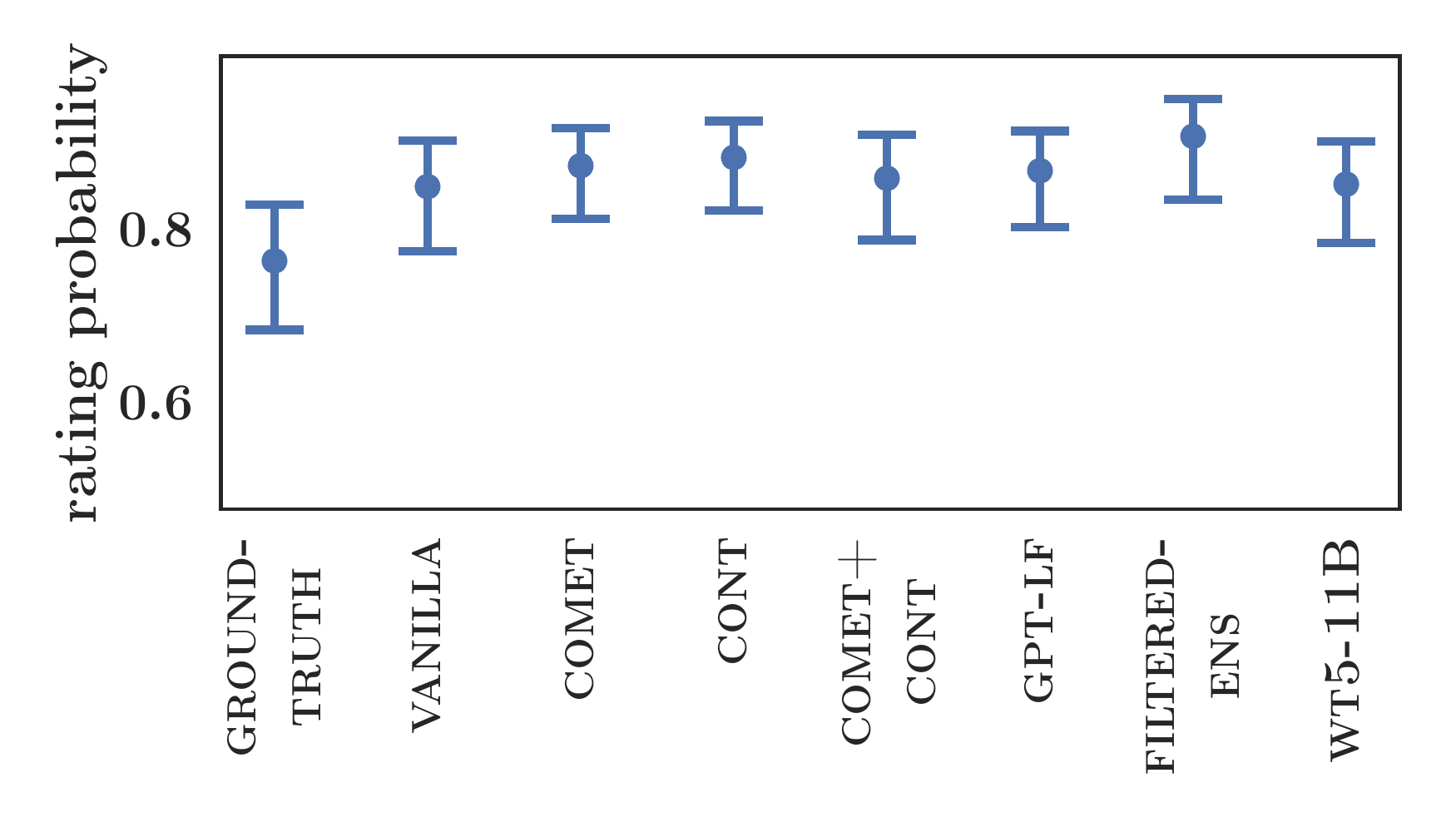}%
        \caption{Label correctness.}\label{fig:effect_display_label}
    \end{subfigure}%
    \begin{subfigure}[t]{.5\textwidth}
        \centering
        \includegraphics[width=\textwidth]{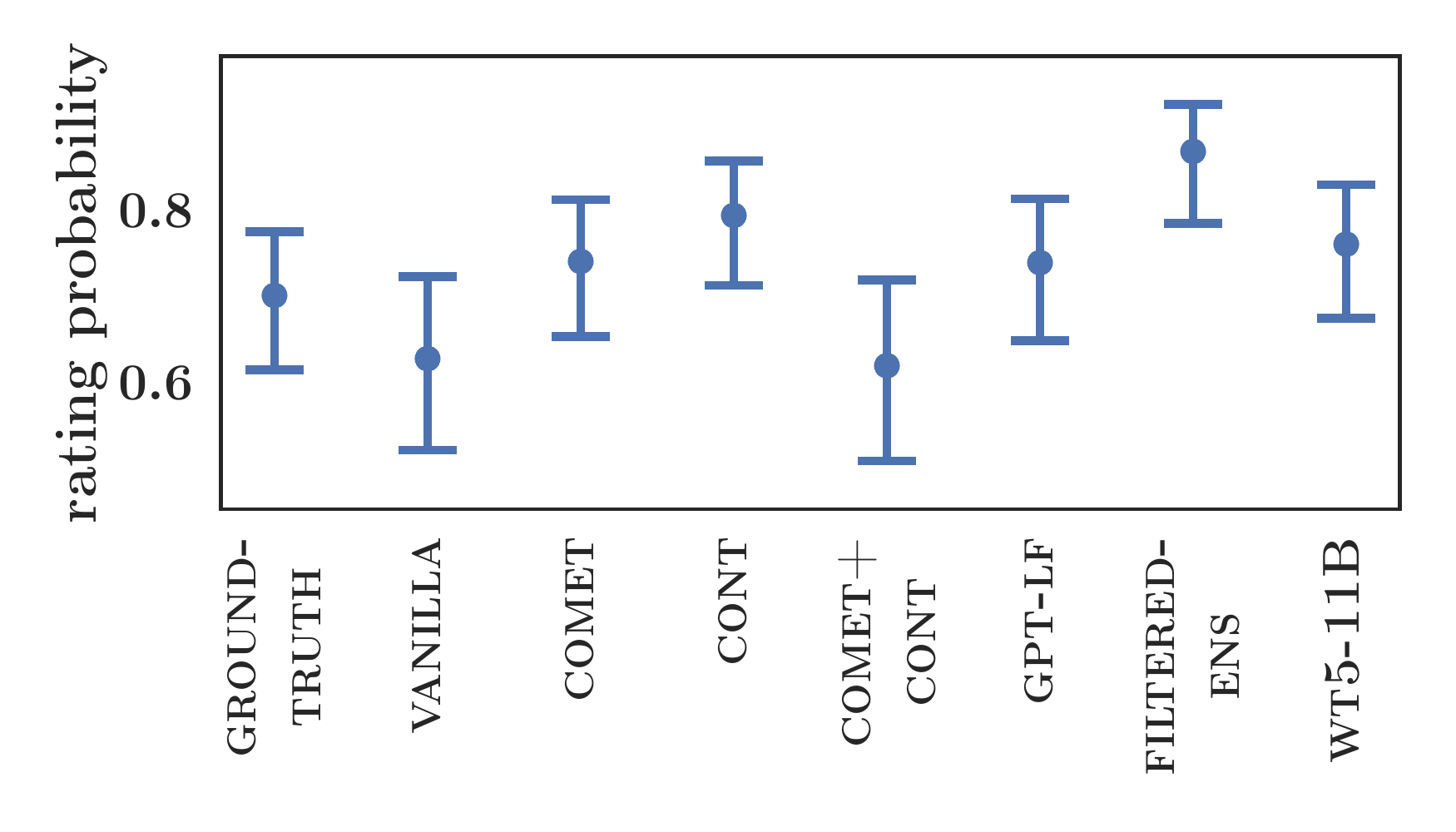}%
        \caption{Explanation correctness.}
    \end{subfigure}
    \begin{subfigure}[t]{.5\textwidth}
        \centering
        \includegraphics[width=\textwidth]{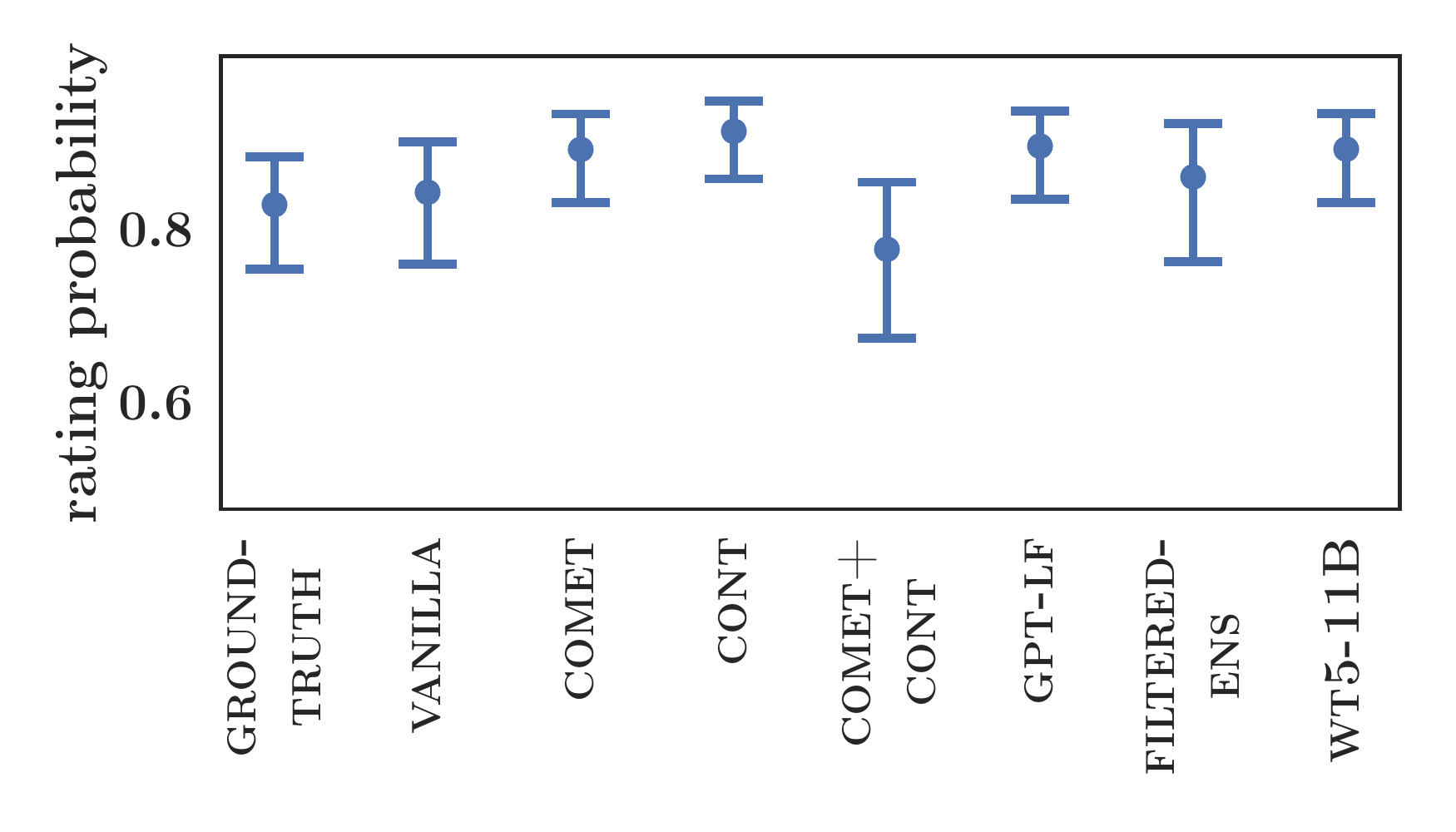}
        \caption{Grammatical correctness.}
    \end{subfigure}%
    \begin{subfigure}[t]{.5\textwidth}
        \centering
        \includegraphics[width=\textwidth]{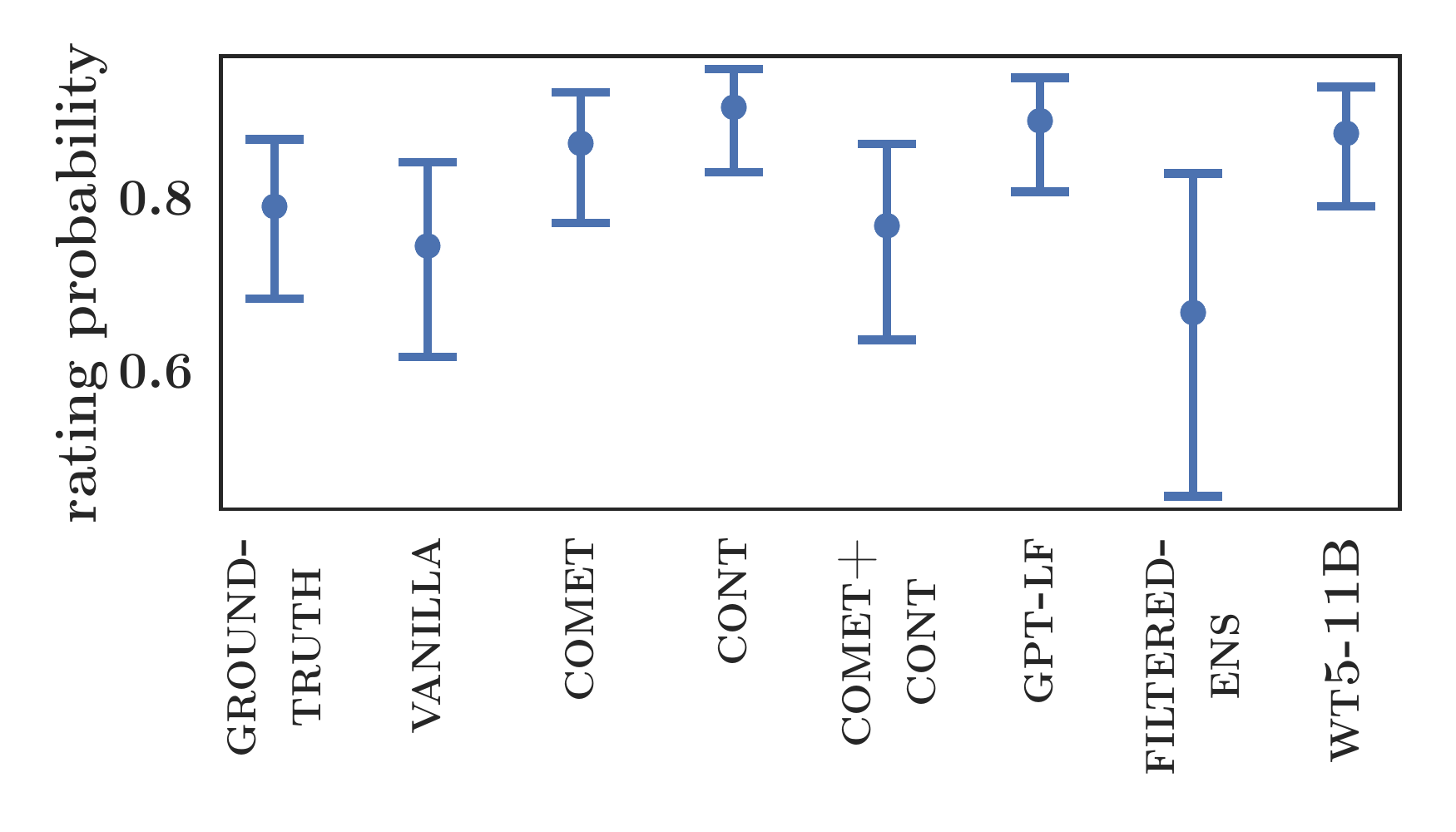}%
        \caption{Commonsense correctness.}
    \end{subfigure}%
    \caption{Effect displays for user ratings of label, explanation, grammatical and commonsense correctness depending on \textit{model type} following \citet{fox_effect_2003}. The rating probability is the probabality that a prediction of a respective model type is perceived to be correct by a human considering fixed effects. Error bars mark 95\% confidence limits.}\label{fig:effect_displays}
\end{figure*}

\paragraph{Label Correctness.}
We do not observe a significant main effect of \textit{model type} ($\chi^2(7)=13.00$, $p=0.0723$) but a significant main effect of \textit{commonsense level}
($\beta=0.28$, $\chi^2(1)=4.54$, $p<0.0331$).\footnote{$\beta$ refers to the estimate of a \textit{high} commonsense level.}

\paragraph{Explanation Correctness.}
We observe a main effect of \textit{model type} ($\chi^2(7)$=24.06, p$<$0.0012) and \textit{commonsense level} ($\beta=0.27$, $\chi^2(1)=7.79$, $p<0.0053$).
For \textit{model type}, a post-hoc Tukey test showed significant differences between \textsc{filtered-ens} and \textsc{vanilla} ($p<0.0055$) as well as \textsc{filtered-ens} and \textsc{comet+cont} ($p<0.0029$).

\paragraph{Grammatical Correctness.}
We observe a main effect of \textit{model type} ($\chi^2(7)=14.20$, $p<0.0479$).
However, a post-hoc Tukey test did not reveal significant differences between any model type pair.
No significant main effect of \textit{commonsense level} was observed
($\beta=0.02$, $\chi^2(1)=0.02$, $p=0.8803$).

\paragraph{Commonsense Correctness.}
We observe a main effect of \textit{model type} ($\chi^2(7)=20.63$, $p<0.0044$).
However, a post-hoc Tukey test did not reveal significant differences between any model type pair.
No significant main effect of \textit{commonsense level} was observed ($\beta=0.07$, $\chi^2(1)=0.25$, $p=0.6163$).

Overall, these results show surprisingly few significant differences between the different model types and conflict with the large differences within automatic evaluation scores.
\section{Discussion}
\paragraph{R1: Effect of External Knowledge.}
We showed that external knowledge can increase label accuracies on e-SNLI as well as on the stress tests.
In addition, we found external knowledge to increase BLEU/BLEURT scores and thus help explanation generation in terms of automatic evaluation.

\paragraph{R2: Implicit Knowledge in Language Models.}
While language models achieve the best scores on general e-SNLI performance, the stress tests showed that they do not succeed in all reasoning types. Thus, for choosing the best way of integrating commonsense knowledge, the final reasoning goal of the model needs to be considered.

\paragraph{R3: Perceived Explanation Quality by Humans.}
We expected the large differences in e-SNLI label accuracy (up to 3.23\%), BLEU (up to 10.17) and BLEURT (0.31) to reflect in human ratings, but none of these maximal differences in scores leads to a significantly different rating for any dependent variable.
Regarding the observed significant differences, \textsc{filtered-ens} is not the best model included in the study with respect to e-SNLI (\textsc{wt5-11B} reaches distinctly higher values for all scores) and, similarly, neither \textsc{vanilla} nor \textsc{comet+cont} are the worst models on any score in Table~\ref{tab:automatic_scores}.
Thus, large accuracy gains do not necessarily imply better models when used in real-world applications with users. In the following, we will further discuss these results.

\paragraph{Superhuman Model or Noisy Ground Truth?}
It is particularly remarkable that the ground truth ratings do not significantly differ from any other model's ratings.
In fact, the ground truth condition ranks in the lower half across all four rating dimensions and yields the lowest probability of receiving label correctness ratings as shown in Figure~\ref{fig:effect_display_label}.
Similarly, \citet{narang_wt5_2020} note that in their experiment the \textsc{wt5-11b} model reaches a 12\%-higher explanation correctness rating than the ground truths.
This indicates that e-SNLI might not be suitable to distinguish performances of today's high-performing models.
While it remains valuable for training, models should be scored on specifically designed evaluation sets, for example an explainable extension of the NLI stress test dataset. 

\paragraph{Limitations and Future Directions.}
Although we evaluated a total of 11 different model architectures and various different sources of external knowledge,
this clearly does not exhaust all possible knowledge sources or architectures.
While our analysis provides insight into the most common knowledge sources integrated into representative model architectures, future work should confirm our findings for additional sources and architectures.
Although our user study already is the largest and most fine-grained evaluation of explainable NLI, future work should further expand the set of dependent variables to potentially reveal effects that are not visible through the lens of our experimental setup.
While our work addresses the task of explainable NLI, we expect that the observed disconnect between automatic and human evaluation applies to further tasks and requires to re-assess model evaluation across explainability tasks.
\section{Conclusion}
In this paper, we addressed three research questions: whether integrating external knowledge can improve explainability for NLI, how effective knowledge implicitly stored in language models is for reasoning, and how humans perceive explanation quality of state-of-the-art natural language inference models.
To answer these questions, we proposed different methods of integrating various knowledge sources into deep learning models.
We found that fine-tuned language models reach the highest performance on e-SNLI as well as the highest average accuracy within the NLI stress test evaluation. However, their performance can break down on numerical reasoning and negations.
In addition to automatic evaluation, we conducted a large-scale human crowdsourcing evaluation and found that high differences in accuracy, BLEU or BLEURT scores do not reflect in significant differences in human ratings of explanation correctness, commonsense inclusion, grammar or label correctness.
This highlights an alarming disconnect between automatic evaluation scores and human ratings, that puts the real-world utility of recent model improvements into question and requires to re-think automatic evaluation across the field of explainable AI.
\section*{Acknowledgement}
We thank the members of the BCAI NLP\&KRR research group and the anonymous reviewers for their helpful comments.
Ngoc Thang Vu is funded by Carl Zeiss Foundation.
\bibliography{anthology,custom}
\bibliographystyle{acl_natbib}
\appendix
\section{Ensemble Architectures}
Figure~\ref{fig:ensemble_extended} depicts the ensemble architectures discussed in Section~\ref{sec:combined_models}.

\begin{figure*}
    \centering
    \includegraphics[width=\textwidth]{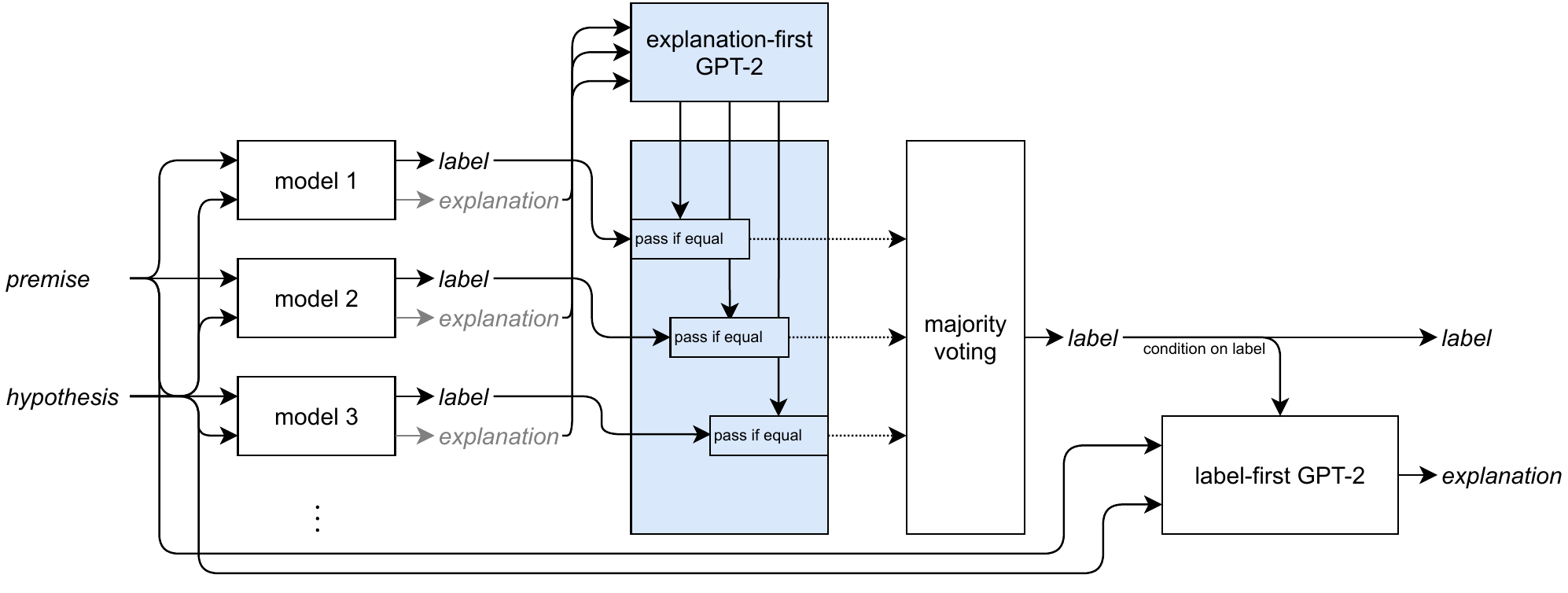}
    \caption{Ensemble architectures. The blue boxes show the consistency-filter extension.}\label{fig:ensemble_extended}
\end{figure*}
        
\section{Study Interface}
Figure~\ref{fig:study_screenshot} shows an example of the study interface used to collect human ratings as discussed in Section~\ref{sec:human_eval}.
\begin{figure*}
    \centering
    \includegraphics[width=0.9\textwidth]{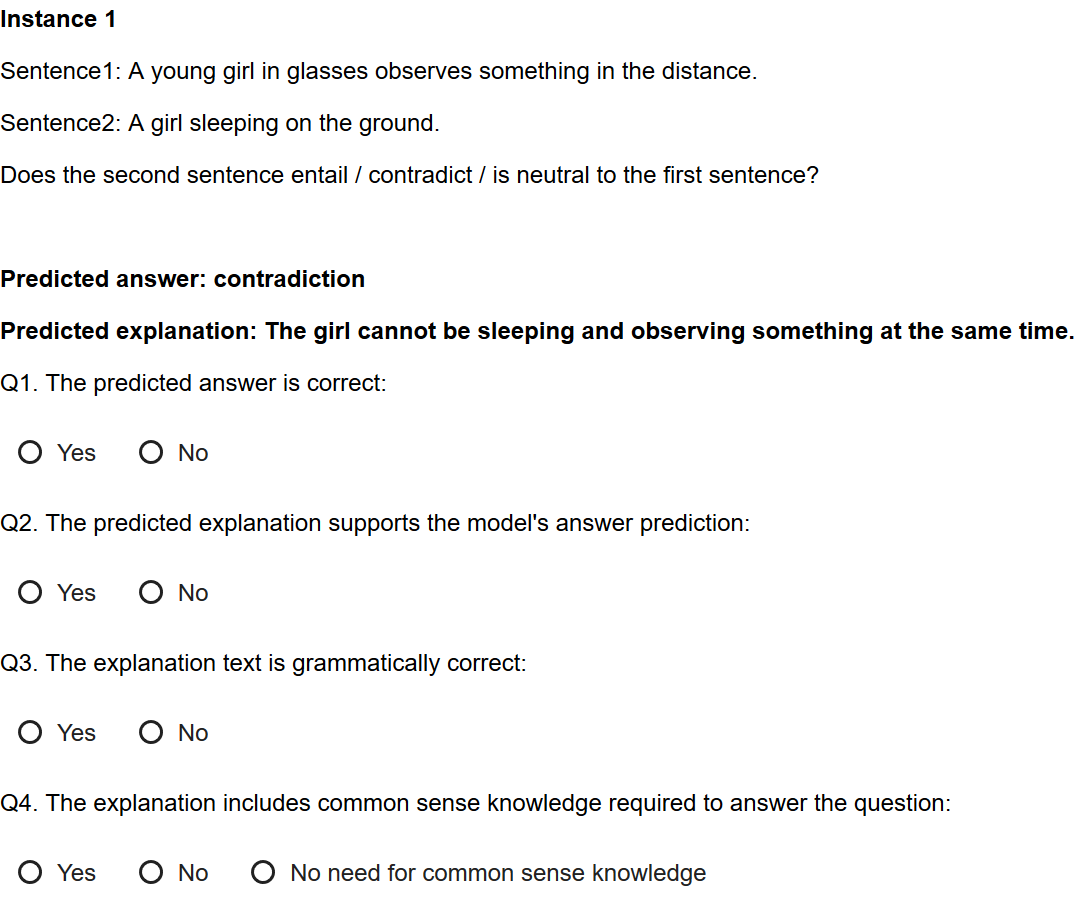}
    \caption{Screenshot of the study interface presented to crowdworkers on Mechanical Turk.}\label{fig:study_screenshot}
\end{figure*}

\section{Knowledge Requirement Annotation}
Table~\ref{tab:annotation_guidelines} lists the annotation guidelines used to decide on low/high levels of required external knowledge as discussed in Section~\ref{sec:study_design}.
Table~\ref{tab:low_high_examples} shows example annotations.
\begin{table*}
\centering
		\begin{tabular}{p{0.04\textwidth}p{0.25\textwidth}p{0.6\textwidth}} 
			\toprule
			\multirow{16}{*}{\rotatebox[origin=c]{90}{Low Level}}			&\multirow{4}{*}{\rotatebox[origin=c]{0}{Pattern Matching}} & The entailment can be decided by matching identical parts in the premise and the hypothesis. \\
			& &Premise:\textit{ A water scene with a sunset in the background.}\\ & &Hypothesis:\textit{There is a water scene with the sunset in the back.} \\ 
			\cmidrule{2-3}
			&\multirow{4}{*}{\rotatebox[origin=c]{0}{Unrelated Negation}} & The entailment can be decided by identifying an unrelated negation. \\
			& &Premise:\textit{ Children bathe in water from large drums.} \\ 
			& &Hypothesis:\textit{ The kids are not reading.} \\ 
			\cmidrule{2-3}
			&\multirow{5}{*}{\rotatebox[origin=c]{0}{Rephrasing}}&The entailment can be decided by simple rephrasing (e.g. replacing a word with a synonym). \\
  			& &Premise:\textit{ A boy dressed in an orange shirt and a helmet is riding a dirt bike in the woods.}\\ 
  			& &Hypothesis:\textit{ A boy in orange rides his dirt bike.} \\ 
  			\cmidrule{2-3}
  			&\multirow{4}{*}{\parbox{0.25\columnwidth}{Easily-Distinguishable Concepts}}&The entailment can be decided by identifying unrelated concepts that have no semantic relation. \\
  			& &Premise:\textit{ Firefighters in full gear are walking up a ladder.} \\ 
  			& &Hypothesis:\textit{ The firefighters are eating lunch.}\\ \midrule
  			\multirow{9}{*}{\rotatebox[origin=c]{90}{High Level}} &\multirow{5}{*}{\rotatebox[origin=c]{0}{Complex Reasoning}}&The entailment can be decided by resolving more complex relations and reasoning using common sense knowledge. \\
  			& &Premise:\textit{ Soccer players are playing a night game and the ball is in the air, while the two teams fight for it. }\\ 
  			& &Hypothesis:\textit{ The sun was shining during the soccer match.}\\
  			\cmidrule{2-3}
  			&\multirow{5}{*}{\rotatebox[origin=c]{0}{Abstract Concepts}}&The entailment can be decided using common sense knowledge about abstractions of concepts.\\
  			& &Premise:\textit{ A girl reaches up to kiss a cat, which is sitting on the counter.} \\ 
  			& &Hypothesis:\textit{ A girl is showing affection towards a cat}.\\ \bottomrule
	\end{tabular}
    \caption{Annotation guidelines used during the annotation of low/high levels of required external knowledge with examples.} \label{tab:annotation_guidelines}
	\end{table*}
\begin{table*}[h]
\centering
		\begin{tabular}{p{0.05\textwidth}l} 
			\toprule
			\multirow{10}{*}{\rotatebox[origin=c]{90}{Low Level}} & Premise: There is a group of children getting their picture taken with presents.\\
			& Hypothesis: Two men carry a Christmas tree.\\
			\cmidrule{2-2}
			& Premise: A woman looks at a plate filled with steam.
\\
			& Hypothesis: The woman is out shopping at the mall.
\\
			\cmidrule{2-2}
			& Premise: Man sitting on bench with a suitcase in front of PADDINGTON sign.
\\
			& Hypothesis: A man sitting with a sign.
\\
			\cmidrule{2-2}
			& Premise: A man grilling a hamburger.
\\
			& Hypothesis: The man is swimming at the bottom of the ocean.
\\
			\cmidrule{2-2}
			& Premise: The African American man protests against unlawful sex.
\\
			& Hypothesis: The man protests.
\\
			\midrule
  			\multirow{10}{*}{\rotatebox[origin=c]{90}{High Level}} & Premise: A boy in a red jacket and black hat sliding on his knees down a snowy hill\\
			& Hypothesis: A child is playing outside.\\
			\cmidrule{2-2}
			& Premise: A man playing a piano.\\
			& Hypothesis: The man's hands are on the keys of a piano.\\
			\cmidrule{2-2}
			& Premise: 3 girls chatting and laughing on the stairwell.\\
			& Hypothesis: Girls are not having a good time.\\
			\cmidrule{2-2}
			& Premise: A man visiting a friend in the hospital.\\
			& Hypothesis: A man and a patient in a hospital room.\\
			\cmidrule{2-2}
			& Premise: Two girls pose along a tree-lined path and blow kisses towards the camera.\\
			& Hypothesis: Two girls are taking pictures outside.\\
			\bottomrule
	\end{tabular}
    \caption{Pairs from the low/high external knowledge requirement annotations sampled from pairs for which annotators agreed.} \label{tab:low_high_examples}
	\end{table*}

\section{Examples of Human Ratings}
Table~\ref{tab:rating_examples} displays various model prediction examples and corresponding examples of human ratings.

\begin{landscape}
\vfill
\begin{table*}[h]
\hspace*{-8cm}
\resizebox{\textheight}{!}{%
\begin{minipage}{\textheight}
\centering
\small
 \begin{tabular}{p{4.5cm}p{2.5cm}p{1.5cm}p{1.5cm}p{1.5cm}p{3.5cm}p{1.cm}p{1.cm}p{1.cm}p{1.cm}}
\toprule                                                                                       Premise &                                           Hypothesis & Common Sense &         Model &     Predicted Label &                                                                                                                              Predicted Explanation &    \multicolumn{4}{c}{Human Ratings} \\
\cmidrule{7-10}
& & & & & & Label &     Expl. &  Grammar & Common Sense \\
\midrule
By a table with several boxes and two coffee cups on it, a woman in a green sweater is opening a small box and another person (whose head cannot be seen) is operating a small handheld electronic device. &           Members of congress opening care packages. &  high &               \textsc{wt5-11b} &        neutral &                                                                Not all women are members of congress. Not all boxes are care packages. &  \cmark  &  \cmark  &  \cmark  &         \cmark \\
\midrule
In front is a guy singing and playing guitar and a white shirted drummer in the back. &                    Some people perform at a concert. &  high &          \textsc{gpt-lf} &        neutral &  Just because a guy singing and playing guitar and a white shirted drummer in the back does not mean they are performing at a concert. &  \cmark  &  \cmark  &  \cmark  &         \cmark \\
\midrule
A man in a black coat and hat and a woman in a blue coat with pink hat snowshoeing on the side of a road with a stoplight in the background . &            A man and woman are outside in the snow . &  high &  \textsc{comet +cont} &     entailment &                                                                                                                 in the snow is outside &  \cmark  &  \cmark  &  \cmark  &         \cmark \\
\midrule
Two individuals dressed up like animals are posing for the camera . &        Two young boys dressed like cows are posing . &   low &  \textsc{comet +cont} &        neutral &                                                                                                               not all people are young &    \xmark &    \xmark &  \cmark  &         \cmark \\
\midrule
A man playing a piano &                                                                        The man's hands are on the keys of a piano. &  high &           \textsc{vanilla} &     entailment &                                                                                                                                                                                                  playing a piano requires one to be on the hands &  \cmark &  \cmark &    \xmark &         \cmark \\

\midrule
Two women are hugging on a path through a grassy area with a cow visible past them. &              There are women outside in this picture &  high &               \textsc{wt5-11b} &     entailment &                                                                                                              A grassy area is outside. &  \cmark  &  \cmark  &  \cmark  &         \cmark \\
\midrule
A woman in a spaghetti strap tank wearing a flower in her hair is staring towards the right . &                                        The woman is sleeping on a pillow . &   low &  \textsc{comet +cont} &  contradiction &                                                                        the woman can not be looking towards a sleeping she is sleeping &  \cmark &    \xmark &    \xmark &         \cmark \\
\bottomrule
\end{tabular}
\centering
\begin{minipage}{22cm}
  \caption{Examples of model predictions and human ratings of the predicted labels and explanations for various model types.}\label{tab:rating_examples}
  \end{minipage}
\end{minipage}
}
\end{table*}

\vfill
\end{landscape}
\end{document}